# Sentiment Analysis of Code-Mixed Languages leveraging Resource Rich Languages

Nurendra Choudhary*, Rajat Singh*, Ishita Bindlish and Manish Shrivastava

Language Technologies Research Centre (LTRC)
Kohli Center on Intelligent Systems (KCIS)
International Institute of Information Technology, Hyderabad, India
`{nurendra.choudhary,rajat.singh}@research.iiit.ac.in`
`ishita.bindlish@students.iiit.ac.in`
`m.shrivastava@iiit.ac.in`

**Abstract** Code-mixed data is an important challenge of natural language processing because its characteristics completely vary from the traditional structures of standard languages.

In this paper, we propose a novel approach called *Sentiment Analysis of Code-Mixed Text (SACMT)* to classify sentences into their corresponding sentiment - positive, negative or neutral, using contrastive learning. We utilize the shared parameters of siamese networks to map the sentences of code-mixed and standard languages to a common sentiment space. Also, we introduce a basic clustering based preprocessing method to capture variations of code-mixed transliterated words. Our experiments reveal that SACMT outperforms the state-of-the-art approaches in sentiment analysis for code-mixed text by 7.6% in accuracy and 10.1% in F-score.

## 1 Introduction

Multilingual societies with decent amount of internet penetration widely adopted social media platforms. This led to the proliferation in usage of code-mixed text. Sentiment analysis of code-mixed data on social media platforms enables scrutiny of political campaigns, product reviews, advertisements and other social trends.

Code-mixed text adopts the vocabulary and grammar of multiple languages and often forms new structures based on its users. This is challenging for sentiment analysis as traditional semantic analysis approaches do not capture meaning of the sentences. Scarcity of annotated data available for sentiment analysis also limit the advances in the field.

In this paper, we aim to solve the limitations and challenges by utilizing a novel unified framework called *"Sentiment Analysis of Code-Mixed Text (SACMT)"*. SACMT model consists of twin Bi-directional Long Short Term Memory Recurrent Neural Networks (BiLSTM RNN) with shared parameters and a contrastive energy function, based on a similarity metric on top. The energy function suits discriminative training for energy-Based models [8].

---

* These authors have contributed equally to this work.

SACMT learns the shared model parameters and the similarity metric by minimizing the energy function connecting the twin networks. Parameter sharing and the Similarity Metric guarantee that, if the sentiment of sentences on both the individual Bi-LSTM networks are same, then they are nearer to each other in the sentiment space, else they are farther from each other. Hence, the representation of *India match jit gayi* (India won the match) and *Diwali ki shubh kamnaye sabko* (Happy Diwali to everybody) are closer to each other and *India match jit gayi* (India won the match) and *Bhai ki movie flop gayi* (Bhai's movie was a flop) are distant from each other. The learned similarity metric models the sentiment similarity of sentences into a common sentiment space.

Transliteration of phonetic languages, like Hindi, into roman script creates several variations of the same word. For example, *"बहुत"*(more) can be transliterated as *bahut,bohot* or *bohut.* To solve this challenge, we perform a preprocessing step that aims at clustering multiple word variations together using a empirical similarity metric.

The rest of the paper is organized as follows. Section 2 describes the previous approaches in the field. Section 3 demonstrates the datasets. Section 4 explain the architecture of SACMT. Section 5 defines the baselines. Section 6 and 7 present the experimental set-up and results respectively. Finally, Section 8 concludes the paper.

## 2   Related Work

Distributional semantics [10] approach captures the words' semantics, but loses out on the information of their sequence in the sentence. Another limitation of the technique is that it considers a word immutable. Hence, it is unable to handle spelling errors, out of vocabulary words properly. [12] assigns polarity scores to individual words. The overall sentiment score of the constituent words assigns the sentence's polarity. Thus, the semantic relation and words' sequence is lost and this leads to incorrect classification. N-grams limit this problem but do not eliminate it completely.

Another line of research, [7], utilizes character level LSTMs to learn sub word level information of social media text. This information then classifies the sentences using an annotated corpus. The model presents an effective approach for embedding sentences. However, the limitation in the approach here is the requirement of abundant data.

### 2.1   Siamese Networks

Siamese networks (shown in figure 1) help in the contrastive learning of a similarity metric without an extensive dependence on the features of the input. [3] introduced siamese networks to solve the problem of signature verification.

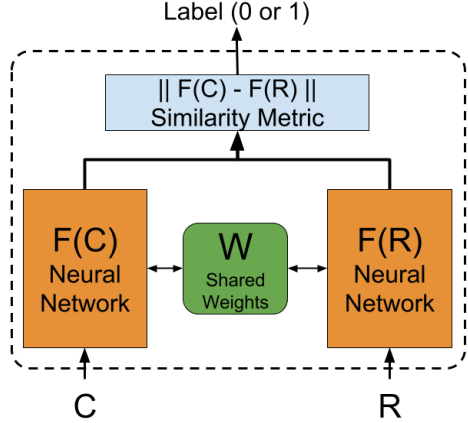

Figure 1: Siamese Network

Later, [4] used the architecture with discriminative loss function for face verification. These networks also effectively enhance the quality of visual search [9,6]. Recently, [5] applied these networks to solve the problem of community question answering.

Let, $F(X)$ be the family of functions with parameters $W$. $F(X)$ is differentiable with respect to $W$. Siamese network seeks a value of the parameter $W$ such that the symmetric similarity metric is small if $X1$ and $X2$ belong to the same category, and large if they belong to different categories. The scalar energy function $S(C, R)$ that measures the sentiments' relatedness between tweets of code-mixed ($C$) text and resource-rich ($R$) language can be defined as:

$$S(C, R) = ||F(C) - F(R)|| \tag{1}$$

In SACMT, we input the tweets from both the languages to the network. We minimize the loss function such that $S(C, R)$ is small if the $C$ and $R$ carry the same sentiment and large otherwise.

| Datasets | Words | Char-trigrams | Positive | Neutral | Negative |
|---|---|---|---|---|---|
| HECM | 43725 | 12842 | 35% | 50% | 15% |
| English-Twitter | 337913 | 197649 | 28% | 46% | 26% |
| English-SemEval'13 | 97280 | 52011 | 40% | 40% | 20% |

Table 1: Properties of the datasets.

## 3   Dataset

We utilize the datasets for testing the architecture on both code-mixed data (Hindi-English) and social media text of a standard language (English). Following are the datasets we considered in our experiments.

- **Hindi-English Code-Mixed (HECM):** The dataset, proposed in [7], consists of 3879 annotated Hindi-English Code-Mixed sentences.
- **English - Twitter:** The dataset, proposed in [11], consists of 103035 annotated English tweets.
- **SemEval 2013:** The dataset, used for SemEval 2013 Task 2B[1], consists of 11338 annotated English tweets.

All the datasets are annotated with three classes - positive, negative and neutral. Table 1 demonstrates the distribution of classes in the above datasets.

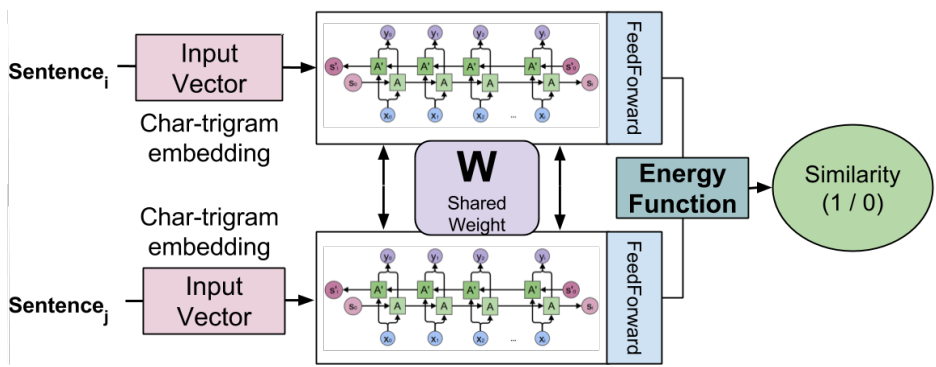

Figure 2: Architecture of SACMT

## 4   Architecture of SACMT

As illustrated in figure 2, SACMT consists of a siamese network with twin character level Bi-LSTM networks with a fully connected layer on top. Bi-LSTMs project sentences on the two ends to a common sentiment space. We connect the yielded sentiment vectors to a layer that measures the similarity between them. The contrastive loss function combines the similarity measure and the label. Back-propagation through time computes the loss function's gradient with respect to the weights and biases shared by the sub-networks.

---

[1] `https://www.cs.york.ac.uk/semeval-2013/task2/index.html`

| Standard | Meaning | Consonants | Captured Variations | | | |
|---|---|---|---|---|---|---|
| खूबसूरत | beautiful | **khbsrt** | **khoobsurat** | khubsurat | khubsoorat | khbsurt |
| क्यूंकि | because | **kynk** | **kyunki** | kiyunki | kiyunkee | kyunkee |
| मेहरबानी | clemency | **mhrbn** | **meherbani** | meharbaani | meharbani | meherbanee |
| आपका | yours | **pk** | **aapka** | apkaa | apka | apkA |

Table 2: Some example variations of standard Hindi words with their replacement shown in bold.

### 4.1 Handling Code-Mixed Word Variations

Transliteration from languages with phonetic script(like Hindi) leads to variation in word depending on the user. We solve this issue using clustering of skip-gram vectors[10]. Skip-gram vectors give the representation of a word in the semantic space based on their context. The variations belong to the same word with similar function implying a similar context. Also, the consonants of these variations in the cases are same (shown in table 2). Hence, we cluster the words based on a similarity metric that captures both these properties. The similarity metric is formally defined below:

$$f(v1, v2) = \begin{cases} sim(vec(v1), vec(v2)) & \text{if v1,v2 have same consonants} \\ 0 & \text{else} \end{cases} \quad (2)$$

where $v1$ and $v2$ are the two variations, $sim$ is a similarity function (like cosine similarity), $vec(v)$ returns the skip-gram vector of $v$ and $f(v1, v2)$ represents the overall similarity between $v1$ and $v2$.

This metric gives us the closest variations for the given word. They together form a cluster and the most frequent word replaces all the other words of the cluster. Here, we assume that the word with the highest frequency also has the most probability of being the correct one.

### 4.2 Primary Embeddings

Code-mixed text, being informal, has challenges such as spelling errors and out of vocabulary words. These variations cannot be dismissed as errors because they capture sentiment. For example, *"Heeey"* conveys positive sentiment, whereas *"Hey"* conveys a neutral sentiment. Hence, we treat character trigrams as immutable units instead of words. This also reduces the computational complexity as the number of words exceeds character trigrams (shown in Table 1).

We input a pair of character based term vectors of the tweet and a label to the twin networks of SACMT. The label indicates whether the samples are nearer or farther to each other in the sentiment space. For positive samples (nearer in the sentiment space), twin networks are fed with tweets' vectors with the same sentiment tags. For negative samples (far away in the sentiment space), twin networks are fed with vectors of tweets with different sentiment tags.

### 4.3 Bidirectional LSTM Network

Each sentence-pair maps into a pair $(a_i, a_j)$ such that $a_i, a_j \in \mathbb{R}^n$ where n is the number of character trigrams in the data.

Bidirectional LSTM [1] model encodes the sequence twice, once forward (original) and once backward (reverse). Back Propagation through Time (BPTT) [2] calculates the weights for both the traversals independently. We apply element-wise Rectified Linear Unit (ReLU) to the output encoding of the BiLSTM. ReLU is defined as: $f(x) = max(0, x)$. The choice of ReLU simplifies back-propagation, causes faster learning and avoids saturation. The architecture's final fully connected layer converts the output of the ReLU layer into a fixed length vector $s \in \mathbb{R}^d$. In our architecture, we have empirically set the value of $d$ to 128. The overall model is formalized as:

$$s = max\{0, W[fw, bw] + b\} \tag{3}$$

where $W$ is a learned parameter matrix (weights), $fw$ is the forward LSTM encoding of the sentence, $bw$ is the backward LSTM encoding of the sentence, and $b$ is a bias term, then passed through an element-wise ReLU.

### 4.4 Training Step

SACMT differs from the other deep learning counterparts due to its property of parameter sharing, which ensures that both the sentences project into the same sentiment space. Given an input $a_i, a_j$ which are embeddings of tweets and a label $y_i \in \{-1, 1\}$, the loss function is defined as:

$$loss(a_i, a_j) = \begin{cases} 1 - cos(a_i, a_j), & \text{if } y = 1; \\ max(0, cos(a_i, a_j) - m), & \text{if } y = -1; \end{cases} \tag{4}$$

where $m$ is the margin by which dissimilar pairs should be moved away. It varies between 0 to 1. The loss function is minimized such that pair of tweets with label 1 (same sentiment) are projected nearer to each other and those with label -1 (different sentiment) are projected farther from each other. The model is trained by minimizing the overall loss function in a batch. The objective is to minimize:

$$L(\Lambda) = \sum_{(a_i, a_j) \in C \cup C'} loss(a_i, a_j) \tag{5}$$

where $C$ contains batch of pairs with same sentiment and $C'$ contains batch of pairs with different sentiment. Back-propagation through time (BPTT) updates the parameters shared by the Bi-LSTM sub-networks.

## 5    Baselines

Following are the baselines defined according to relevant previous approaches.

– **Average Skip-gram Vectors (ASV):** Word2Vec [10] provides a vector for each word. We average the words' vectors to get the sentence's vector. So, each sentence vector is defined as:

$$V_s = \frac{\sum_{w \in W_s} V_w}{|W_s|} \tag{6}$$

where $V_s$ is the vector of the sentence $s$, $W_s$ is the set of the words in the sentence and $V_w$ is the vector of the word $w$.

After obtaining each message's embedding, we train a L2-regularized logistic regression (with $\epsilon$ equal to 0.001).

– **Subword LSTM (SWLSTM):** We take the approach, proposed in [7], as the baseline for Hindi-English Code-Mixed data. Character embeddings of the sentence are input and Convolutional Neural Networks capture sub-word level information from the sentence. These embeddings of the tweets classification into different sentiment classes.

## 6    Experiments

We conduct different experiments to compare the model with diverse inputs and also against the previous approaches in the field. The first experiment (section 6.1) analyzes the performance of SACMT on varying language pairs. In the second experiment (section 6.2), we compare SACMT against the baselines defined in section 5. The third experiment (section 6.3) tests the added performance boost due to the preprocessing step that handles variations. In the final experiment (section 6.4), we provide an extension based on emojis retrieved from social media instead of sentiment tags.

| Models | Accuracy | Precision | Recall | F-score |
|---|---|---|---|---|
| SNASA(HE-HE) | 71.3% | 0.693 | 0.668 | 0.680 |
| SACMT(HE-Eng) | 77.3% | 0.770 | 0.749 | 0.759 |
| SACMT(Eng-Eng) | **79.8%** | **0.795** | **0.763** | **0.778** |

Table 3: Comparison of SACMT trained on different language pairs.

### 6.1    Experiments for different language pairs

The experiment is a classification task. We consider the Hindi-English Code-Mixed (HECM) sentences and align them with the English sentences from the Twitter datasets of the same sentiment and label them 1 (positive samples). Likewise, we also randomly sample equal number of English sentences with different sentiment (negative samples) and label them -1. We use this model (SACMT(HE-Eng)) to observe the advantages of training Hindi-English Code-Mixed data in conjunction with English sentences.

Also, we construct the input data by aligning each HECM sentence with corresponding HECM sentences of the same sentiment (positive samples) and label them 1. Likewise, we randomly sample equal number of HECM sentences with different sentiment (negative sample) and label them -1. Same method constructs the model for English sentences from Twitter dataset. We create these models (SACMT(HE-HE) and SACMT(Eng-Eng)) to observe the advantages that shared parameters of siamese network provide in overall sentiment analysis.

Table 3 demonstrates the performance of these models .

| **Model** | Accuracy | Precision | Recall | F-score |
|---|---|---|---|---|
| ASV | 57.6% | 0.5132 | 0.5336 | .5232 |
| SWLSTM | 69.7% | 0.646 | 0.671 | 0.658 |
| SACMT(HE-HE) | 71.3% | 0.68 | 0.665 | 0.672 |
| SACMT(HE-Eng) | **77.3%** | **0.766** | **0.753** | **0.759** |
| Improvement | 7.6% | 0.12 | 0.082 | 0.101 |

Table 4: Comparison of SACMT with the baselines. ASV and SWLSTM denote the Average Skip-gram vector and Sub-Word LSTM model respectively.

### 6.2   Comparison with the baselines

In this experiment, we compare SACMT with the baselines defined in Section 5.

We perform contrastive learning of our model using data made by aligning each HECM sentence with a set of English and HECM positive samples (with the same sentiment) with label 1 and a set of negative samples (with different sentiment) of the same size with label -1. We consider the models SACMT(HE-Eng) and SACMT(HE-HE) for comparison with the baselines.

Both of the above models are evaluated on the HECM dataset. For appropriate comparability, we train and evaluate the baselines on the HECM dataset.

Table 4 demonstrates the performance of baselines and trained models for the experiment.

| **Models** | With Preprocessing | | | | Without Preprocessing | | | |
|---|---|---|---|---|---|---|---|---|
| | Accuracy | Precision | Recall | F-score | Accuracy | Precision | Recall | F-score |
| ASV | 59.7% | 0.5893 | 0.5597 | 0.5741 | 57.6% | 0.5132 | 0.5336 | 0.5232 |
| SWLSTM | 71.2% | 0.669 | 0.692 | 0.680 | 69.7% | 0.646 | 0.671 | 0.658 |
| SNASA(HE-HE) | 72.4% | 0.713 | 0.694 | 0.703 | 71.3% | 0.693 | 0.668 | 0.680 |
| SACMT(HE-Eng) | 78.0% | 0.775 | 0.759 | 0.767 | 77.3% | 0.770 | 0.749 | 0.759 |

Table 5: Difference in performance of SACMT with and without the preprocessing step (handling word variations).

### 6.3  Affect of handling word variations

To analyze the impact of handling word variations on the overall sentiment analysis task. We train all the models defined (including the baselines), both on the regular data and preprocessed data. The difference in the performance is given in Table 5.

| Emojis | Class | Eng | Spa | Hin | Tel |
|--------|-------|-----|-----|-----|-----|
| ❤️ 😎 😁 😊 | Positive | 37% | 36% | 39% | 39% |
| 🙂 🤔 😕 😌 | Neutral | 31% | 30% | 31% | 31% |
| 😫 😟 😭 ❌ | Negative | 32% | 34% | 30% | 30% |

Table 6: Distribution after mapping Emojis to respective sentiment classes.

|  | SNASA | | Emoji-SNASA | |
|--|-------|--|-------------|--|
| Dataset | A(%) | F1 | A(%) | F1 |
| HECM | 71.3% | 0.680 | 74.8% | 0.74 |
| HECM-English | 77.3% | 0.759 | 81.5% | 0.80 |
| English | 79.8% | 0.795 | 82.25% | 0.81 |

Table 7: Performance enhancement due to emojis in sentiment analysis.

### 6.4  Emoji based approach with SACMT (Emoji-SACMT)

In our previous experiment (Section 6.1), we observed that in several test scenarios, limited correlation between the language pair leads to incorrectly classified tweets. Emojis are characters used in social media to communicate context inexpressible by normal characters. A major application of these emojis is expressing sentiment. So, we use the emojis available in our social media datasets to align language pairs instead of sentiment tags. Three annotators manually classify the emojis in the dataset into sentiment classes. We only consider the emojis if all the three annotators are in agreement. The distribution of the formed sentiment classes is given in Table 6.

We align each English sentence with a set of positive samples (with the same emoji) with label 1 and a set of negative samples (with different emoji) of the same size with label -1. The results for the experiment are given in Table 7.

## 7  Evaluation of the Experiments

From the first experiment's results (Table 4), we observe that SACMT(Eng-Eng) outperforms the other language pairs. Eng-Eng has the most number of training samples. This presents the significant impact of the training samples' number on the architecture.

In the second experiment, we observe that SACMT outperforms the state-of-the-art approaches by 7.6% in accuracy and 10.1% in F-score. The additional advantage of shared parameters project the sentences into sentiment space in conjunction with each other. The shared parameters create sentence representations, in accordance to the similarity metric specific to the problem.

Also, we observe that SACMT utilizes language with higher resources to improve the performance of sentiment analysis in the code-mixed text significantly. This allows us to leverage the resources of another language (English in this case) to improve the performance on the code-mixed text.

The third experiment demonstrates the effectiveness of handling word variations. We observe a boost in performance of both the previous approaches and the proposed model by applying a basic preprocessing step.

Multiple times incorrect correlation between the languages in the pair misclassified a sentence. We corrected this behavior by using emojis in twitter dataset to increase the number of usable sentences in establishing correlation. To verify this behavior, we conducted another experiment in section 6.4 to approach this from the perspective of emojis instead of sentiment tags. The experiment's results (given in table 7) demonstrate that emojis lead to better accuracy. This is seen because emojis lead to a better correlation between the pair's languages. However, the drawback of this approach is that emojis do not always represent perfect sentiment and hence will increase the performance only if the data taken has limited noise.

## 8   Conclusions

In this paper, we propose SACMT for sentiment analysis of code-mixed text which solves the problem by using shared parameters to project the sentences into a common sentiment space. SACMT employs twin Bidirectional LSTM networks with shared parameters to capture a sentiment based representation of the sentences. We used these sentiment based representations in conjunction with a similarity metric to group sentences with similar sentiment together.

Experiments conducted on the datasets reveal that SACMT outperforms the state-of-the-art approaches significantly. SACMT leverages the resources of other languages to improve the sentiment analysis' performance on code-mixed text.

The word variations' handling, also further, increased performance of all the trained models, including baselines. An emoji based approach used in conjunction with SACMT boosts the performance of overall sentiment analysis further.

As part of future work, we would like to investigate more tasks solvable using resource rich languages as a leverage.

## References

1. Barbieri, F., Ballesteros, M., Saggion, H.: Are emojis predictable? arXiv preprint arXiv:1702.07285 (2017)
2. Boden, M.: A guide to recurrent neural networks and backpropagation. the Dallas project (2002)
3. Bromley, J., Guyon, I., LeCun, Y., Säckinger, E., Shah, R.: Signature verification using a" siamese" time delay neural network. In: Advances in Neural Information Processing Systems. pp. 737–744 (1994)

4. Chopra, S., Hadsell, R., LeCun, Y.: Learning a similarity metric discriminatively, with application to face verification. In: Computer Vision and Pattern Recognition, 2005. CVPR 2005. IEEE Computer Society Conference on. vol. 1, pp. 539–546. IEEE (2005)

5. Das, A., Yenala, H., Chinnakotla, M., Shrivastava, M.: Together we stand: Siamese networks for similar question retrieval. In: Proceedings of the 54th Annual Meeting of the Association for Computational Linguistics (Volume 1: Long Papers). vol. 1, pp. 378–387 (2016)

6. Ding, S., Cong, G., Lin, C.Y., Zhu, X.: Using conditional random fields to extract contexts and answers of questions from online forums. In: ACL. vol. 8, pp. 710–718 (2008)

7. Joshi, A., Prabhu, A., Shrivastava, M., Varma, V.: Towards sub-word level compositions for sentiment analysis of hindi-english code mixed text. In: COLING. pp. 2482–2491 (2016)

8. LeCun, Y., Huang, F.J.: Loss functions for discriminative training of energy-based models. In: AIStats (2005)

9. Liu, Y., Li, S., Cao, Y., Lin, C.Y., Han, D., Yu, Y.: Understanding and summarizing answers in community-based question answering services. In: Proceedings of the 22nd International Conference on Computational Linguistics-Volume 1. pp. 497–504. Association for Computational Linguistics (2008)

10. Mikolov, T., Sutskever, I., Chen, K., Corrado, G.S., Dean, J.: Distributed representations of words and phrases and their compositionality. In: Advances in neural information processing systems. pp. 3111–3119 (2013)

11. Mozetič, I., Grčar, M., Smailović, J.: Multilingual twitter sentiment classification: The role of human annotators. PloS one 11(5), e0155036 (2016)

12. Mukku, S.S., Choudhary, N., Mamidi, R.: Enhanced sentiment classification of telugu text using ml techniques. In: SAAIP@ IJCAI. pp. 29–34 (2016)